\documentclass[10pt]{article}

\usepackage[utf8]{inputenc}
\usepackage[T1]{fontenc}
\usepackage{times}
\usepackage[margin=1in]{geometry}
\usepackage[round]{natbib}
\usepackage{authblk}
\usepackage{microtype}
\usepackage{amsmath,amssymb}
\usepackage{booktabs}
\usepackage{tabularx}
\usepackage{longtable}
\usepackage{array}
\usepackage{multirow}
\usepackage{xcolor}
\usepackage{graphicx}
\usepackage{tikz}
\usetikzlibrary{arrows.meta,positioning,shapes.geometric,fit,calc}
\usepackage{listings}
\usepackage{hyperref}
\usepackage{xurl}
\usepackage{xspace}
\usepackage{placeins}

\definecolor{CPAblue}{HTML}{245B8F}
\definecolor{CPAteal}{HTML}{167A72}
\definecolor{CPAorange}{HTML}{A85D1A}
\definecolor{CPAlight}{HTML}{F3F7FA}
\definecolor{CPAgray}{HTML}{555555}
\hypersetup{colorlinks=true,linkcolor=CPAblue,citecolor=CPAblue,urlcolor=CPAblue,
  pdftitle={Canonical Procedural Actions: An Auditable Annotation Protocol for Tool-Use Agent Traces},
  pdfauthor={Songqi Li, Dongqing Li, Zheqiao Cheng}}
\newcolumntype{Y}{>{\raggedright\arraybackslash}X}

\newcommand{\cpas}{\textsc{CPAs}\xspace}
\newcommand{\induce}{\textsc{Induce}}
\newcommand{\applymode}{\textsc{Apply}}
\newcommand{\match}{\textsc{MatchExisting}}
\newcommand{\propose}{\textsc{ProposeNew}}
\newcommand{\abstain}{\textsc{Abstain}}
\newcommand{\CPAMacroAgreement}{0.974}
\newcommand{\CPAAuditLower}{0.971}
\newcommand{\CPAAuditUpper}{0.991}
\newcommand{\CPADisagreementAnchors}{16}
\newcommand{\CPACoarseAgreement}{0.986}

\newcommand{\CPAMessageAgreement}{0.971}

\newcommand{\CPAQueryCount}{6}
\newcommand{\CPAQueryTraceCount}{6}
\newcommand{\CPAMultiAnchorCOne}{70}
\newcommand{\CPAMultiAnchorCTwo}{67}
\newcommand{\CPAContextMatched}{395}
\newcommand{\CPAContextAgreement}{0.798}

\title{Canonical Procedural Actions:\\
An Auditable Annotation Protocol for Tool-Use Agent Traces}

\author[1]{Songqi Li\thanks{Corresponding author: \texttt{sl2097@cam.ac.uk}}}
\author[2]{Dongqing Li\thanks{\texttt{dongqing.li@kellogg.ox.ac.uk}}}
\author[3]{Zheqiao Cheng\thanks{\texttt{chengzheqiao@gmail.com}}}
\affil[1]{Cambridge Language Sciences, Raised Faculty Building,
Sidgwick Avenue, Cambridge CB3 9DA, United Kingdom}
\affil[2]{Department of Statistics, University of Oxford,
24--29 St Giles', Oxford OX1 3LB, United Kingdom}
\affil[3]{Zhejiang University, China}

\date{September 2026}

\begin{document}
\maketitle

\begin{abstract}
Tool-use agent traces identify messages and API calls, but procedural analyses
also need explicit units of action and inspectable links to their evidence.
We present Canonical Procedural Actions (\cpas), an annotation protocol that
records a procedural function, its first agent-event anchor, the agent events
that realize it, and separate contextual evidence.  Multiple actions may share
a message anchor without an inferred within-message order.  A retail case
study produces a versioned 24-entry codebook through open induction, recorded
consolidation, and successive application audits.  Two isolated LLM contexts
annotate 32 trajectories disjoint from development at the trajectory level,
producing 499 and 491 occurrences with anchor--label overlap $A=0.982$.
Requiring identical context-event references reduces overlap to $0.798$.
These are structural repeatability measures, not semantic accuracy: 16 of 26 task IDs
also occur in development, and historical tool payloads were truncated to
110 characters.  Retrospective controls show that collapsing all labels raises
overlap to $0.986$, while simple endpoint rules reproduce the tool-anchored
portion with $0.997$ overlap.  Assistant-message actions have $0.971$ overlap,
with a per-label minimum of $0.816$.  Applying the frozen codebook to 244
further trajectories yields 4,058 records, including eight diagnostic outcomes.
The contribution is an explicit, auditable annotation instrument and a case
study of its construction and measurement limits; human-reference validity
and downstream utility remain to be established.
\end{abstract}

\section{Introduction}
\label{sec:introduction}

An agent trace already records who spoke, which tool was called, and what it
returned.  Those structural fields answer many useful questions directly.
Procedural analysis asks an additional question: what observable function did
an event serve in the task?  A retail assistant message may both present
product options and ask the customer to choose one; several retrieval calls
may jointly realize one record-gathering action.  Neither a tool-name sequence
nor a one-label-per-message representation expresses both cases.

We introduce \emph{Canonical Procedural Actions} (\cpas) as an annotation
instrument for such analyses.  Each occurrence records a procedural function,
its first agent-event anchor, the agent events that realize it, and separate
customer/tool/environment events cited as context.  The representation is an
anchor-aware multiset: it retains repeated occurrences and multiple functions
at one message, while declining to infer their within-message order.
Definitions use four fields---proximal goal, principal output, state change,
and preconditions---to make the intended meaning and its boundaries inspectable.
The fields guide annotation judgement; they do not constitute a verified
semantic parser.

Functional and overlapping dialogue annotation already has a substantial
history \citep{bunt2019dialogbank,bunt2020iso}.  Procedural representations of
agent trajectories also support vocabulary induction and behavioural
comparison \citep{oderinwale2026procgrep}.  Our contribution is the specific
combination of occurrence-level procedural definitions, shared anchors,
separate realization/context references, and recorded codebook decisions for
tool-use traces.  We examine how this instrument is constructed and what its
consistency scores can establish, using a retail case study.

The study contributes three inspectable artifacts and analyses:
\begin{enumerate}
  \item An occurrence schema with explicit multiplicity and evidence-reference
  invariants, together with a 24-entry retail codebook and applied instructions.
  \item A construction record from open induction through revision, a frozen
  candidate audit, and application to 244 further trajectories yielding 4,058
  occurrence records.  Diagnostic proposals and abstentions remain visible.
  \item A retrospective audit of the original paired outputs: label support,
  tool/message strata, evidence-reference overlap, task-cluster uncertainty,
  label-coarsening controls, and
  native-event/endpoint-rule baselines.
\end{enumerate}

The final paired audit gives anchor--label overlap $A=0.982$ on 32
trajectory-disjoint traces, but this result requires qualification.  The
underlying tasks partly overlap development; the historical input view clips
tool payloads; and both annotators are isolated LLM contexts without a human
reference.  Collapsing all labels improves overlap, and simple endpoint rules
almost exactly reproduce the tool portion.  These results prevent interpreting
high agreement as evidence of correct labels, useful granularity, or an
improvement over native traces.

CPA makes certain analyses operational: for example, one can retrieve every
message annotated as both presenting variants and requesting mutation input,
then inspect its cited events.  Section~\ref{sec:query-example} demonstrates
that query on the archived outputs.  Whether the annotations make an analyst
more accurate or faster is a separate empirical question.  The present paper
establishes the representation and reports its conditional repeatability;
it does not report a validated failure detector or an agent-performance gain.

\section{Related work}
\label{sec:related}

\paragraph{Agent trajectories as behavioural data.}
$\tau$-bench treats evaluation as an interaction among a user, an agent, and
domain tools, rather than as a final-answer-only exercise
\citep{yao2025taubench}.  This makes a richer empirical question possible:
what did the agent actually do on its way to an outcome?  Raw traces provide
the evidence, but endpoint names and conversational turns are interface-level
objects.  CPA contributes a layer between that evidence and later behavioural
analysis: a small, versioned vocabulary of observable procedural functions
whose source events can still be inspected.

\paragraph{Grounded action annotation.}
The distinction between a purposeful action and the operation through which it
is carried out has a long history in activity theory \citep{leontiev1978}.
Here it is a pragmatic modelling device: an action is defined by its observable
goal, output, state transition, and preconditions, rather than by a particular
tool or phrase.  This is close in spirit to stand-off annotation frameworks,
which preserve a link between labels and the source material that supports them
\citep{birdliberman2001}.  The relevant unit in our data is neither a token nor
a conventional non-overlapping span.  One message can realize several actions,
so the representation must retain multiplicity at a shared anchor.

\paragraph{Dialogue functions and procedural vocabularies.}
ISO dialogue-act annotation distinguishes communicative functions and supports
application-specific extensions \citep{bunt2020iso}.  DialogBank uses functional
segments that may be discontinuous or overlapping \citep{bunt2019dialogbank}.
Thus multiple functions within a turn and source-linked annotation are not new
in themselves.  CPA applies these concerns to procedural units spanning tool
calls and assistant messages, with explicit event anchors, realization/context
separation, and a versioned retail codebook.  This is a domain-specific design
choice rather than a demonstrated advantage over ISO annotation.

ProcGrep induces procedural vocabularies from coding-agent traces, including
BPE-based action chunking, and studies behavioural fingerprints and procedural
queries \citep{oderinwale2026procgrep}.  It is a close precedent for canonical
trajectory representations and their analytical use.  CPA focuses on defining
and auditing semantically named occurrence units before such downstream
analysis, including co-anchored message functions.  We have not compared the
two systems on a common corpus and make no performance-superiority claim.

\paragraph{Codebooks, LLMs, and measurement claims.}
Inductive coding normally involves proposing categories, comparing cases,
refining definitions, and recording decisions \citep{charmaz2006,braunclarke2006}.
LLMs make large-scale structured coding practical
\citep{dai2023llmintheloop,gilardi2023chatgpt}, but they also make it easy to
mistake a fluent label for a validated measurement.  Work on LLM-assisted
annotation emphasizes the need to distinguish model suggestions, human
judgement, and downstream claims \citep{pangakis2025humancentered,
schroeder2025anchoring}.  We therefore treat the library, prompt, and evaluator
as a versioned instrument.  The audit measures whether that instrument yields
reproducible anchored occurrences under isolated runs; it does not replace a
human-reference study.

\paragraph{From action labels to later analysis.}
Agreement metrics must respect the annotation unit being compared
\citep{artsteinpoesio2008}.  Our anchor-level multiplicity measures are chosen
for the frozen CPA schema, rather than borrowed from a token-segmentation
setting.  A provenance-preserving action layer can eventually support
process-mining or event-log analysis \citep{vanderAalst2011processmining,
rebmann2022semantic}.  This paper stops one step earlier: it specifies an annotation layer on which such analyses could be tested.

\section{An auditable CPA representation}
\label{sec:method}

A CPA is designed to make a trajectory claim inspectable.  A reader should be
able to ask: \emph{which agent event realized this action, what other agent
events belong to it, and which observed events establish its meaning?}  The
representation therefore keeps action realization and contextual evidence
separate.

\paragraph{A short vignette.}
Suppose a customer asks whether a product variant can be used to change an
order.  The agent retrieves the order, observes the available variants, then
replies: ``Blue is available; would you like me to make that change?''  The
retrieval tool call is one CPA, while the reply contains two:
\path{PRESENT_VARIANT_OPTIONS} exposes the choice set and
\path{SOLICIT_MUTATION_INPUT} asks the customer to authorize or specify
the pending change.  Both reply clauses share an event anchor.  The tool result
is evidence for their meaning, rather than an action in its own right.  CPA is
a representation that retains those distinctions without asking an annotator to
infer a hidden plan.

\subsection{Occurrence model}

For trajectory $t$, let $E_t=(e_{t,1},\ldots,e_{t,T_t})$ be its ordered event
sequence, let $\mathcal E_t=\{e_{t,1},\ldots,e_{t,T_t}\}$ be its event set,
and let $\mathcal G_t\subseteq\mathcal E_t$ be the agent-generated events.
A CPA is an occurrence record,
\begin{equation}
c_{t,j}=(t,j,h_{t,j},\mathbf a_{t,j},\mathbf x_{t,j},
         \ell_{t,j},d_{t,j},\xi_{t,j}),
\label{eq:occurrence}
\end{equation}
where $\mathbf a_{t,j}=(a_1,\ldots,a_m)$ lists the agent events that jointly
realize the action and $h_{t,j}=a_1$ is its \textsc{ActionAnchor}.
$\mathbf x_{t,j}$ contains the customer, tool-result, or environment events
that establish the action's goal, inputs, output, state transition, or outcome.
The remaining fields hold a library label $\ell_{t,j}$ (or a candidate label),
a decision $d_{t,j}\in\{\match,\propose,\abstain\}$, and compact outcome,
confidence, review, and evidence metadata $\xi_{t,j}$.  The frozen schema
requires
\begin{equation}
\mathbf a_{t,j}\subseteq\mathcal G_t,\qquad
h_{t,j}=\min\mathbf a_{t,j},\qquad
\mathbf x_{t,j}\subseteq\mathcal E_t\setminus\mathcal G_t .
\label{eq:provenance}
\end{equation}
An action can have no contextual evidence IDs when the agent event itself is
sufficient evidence.

\begin{figure}[t]
\centering
\resizebox{0.96\linewidth}{!}{
\begin{tikzpicture}[
  event/.style={draw=CPAgray, rounded corners=1pt, fill=CPAgray!8,
    minimum width=25mm, minimum height=12mm, align=center, font=\scriptsize},
  agent/.style={event, draw=CPAblue, fill=CPAblue!10},
  cpa/.style={draw=CPAteal, rounded corners=2pt, fill=CPAteal!9,
    minimum width=34mm, minimum height=13mm, align=center, font=\scriptsize},
  arrow/.style={-{Stealth[length=1.2mm]}, semithick, draw=CPAgray},
  ctx/.style={-{Stealth[length=1.1mm]}, thin, dashed, draw=CPAorange},
  ctxlab/.style={font=\tiny, text=CPAorange, fill=white, inner sep=1pt}
]
\node[event] (e1) at (0,0) {customer $e_1$\\[-1pt]
  \tiny ``change this order to blue?''};
\node[agent] (e2) at (3.2,0) {agent tool call $e_2$\\[-1pt]
  \tiny retrieve order record};
\node[event] (e3) at (6.4,0) {tool result $e_3$\\[-1pt]
  \tiny order and blue availability};
\node[agent] (e4) at (9.6,0) {assistant message $e_4$\\[-1pt]
  \tiny ``Blue is available; make change?''};

\node[font=\tiny, text=CPAgray] at (0,-0.95) {request};
\node[font=\tiny, text=CPAblue] at (3.2,-0.95) {action event};
\node[font=\tiny, text=CPAgray] at (6.4,-0.95) {context evidence};
\node[font=\tiny, text=CPAblue] at (9.6,-0.95) {two functions};

\node[cpa] (c1) at (3.2,-3.4) {\path{RETRIEVE_ORDER_RECORD}\\
  anchor $e_2$};
\node[cpa] (c2) at (7.7,-3.4) {\path{PRESENT_VARIANT_OPTIONS}\\
  anchor $e_4$};
\node[cpa] (c3) at (11.5,-3.4) {\path{SOLICIT_MUTATION_INPUT}\\
  anchor $e_4$};

\draw[arrow] (e2.south) -- (c1.north);
\draw[arrow] (e4.south) -- (c2.north);
\draw[arrow] (e4.south) -- (c3.north);

\draw[ctx] (e3.south) to[bend right=18]
  node[ctxlab, pos=0.55, above left] {context} (c1.north east);
\draw[ctx] (e3.south) to[bend left=10]
  node[ctxlab, pos=0.55, right] {context} (c2.north west);
\draw[ctx] (e1.north) -- ++(0,0.55) -| ([xshift=9mm]c3.north)
  node[ctxlab, pos=0.25, above] {context};

\node[font=\tiny, text=CPAgray, align=center, text width=70mm]
  at (9.6,-4.7)
  {The two $e_4$ occurrences share an anchor; no arrow orders them.};
\end{tikzpicture}}
\caption{\textbf{CPA provenance.} An occurrence is anchored to the first
agent event that realizes it and cites contextual evidence separately.  The two
right-most CPAs share an assistant-message anchor; their horizontal placement
does not assert an order between clauses.}
\label{fig:anchor-schema}
\end{figure}
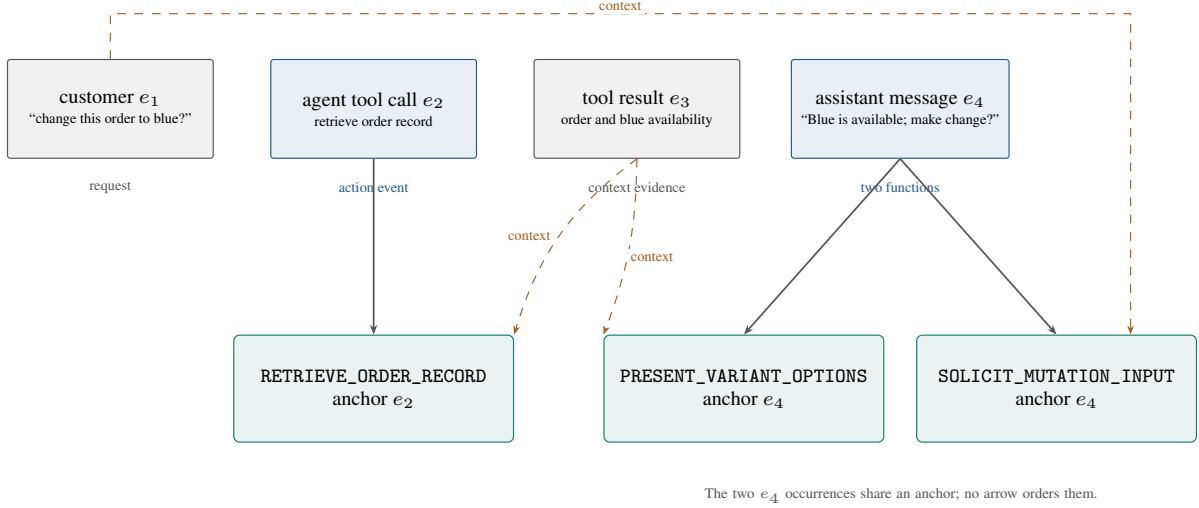

The record is deliberately anchor-aware rather than a partition of events.
If an assistant message presents available variants and asks the customer to
authorize a change, it can yield both \path{PRESENT_VARIANT_OPTIONS} and
\path{SOLICIT_MUTATION_INPUT}.  Formally,
\begin{equation}
h_{t,j}=h_{t,j'}\ \not\!\Longrightarrow\ c_{t,j}=c_{t,j'}.
\label{eq:same-anchor}
\end{equation}
The observed order is consequently partial:
\begin{equation}
c_{t,j}\prec_{\mathrm{obs}}c_{t,j'}
\quad\Longleftrightarrow\quad h_{t,j}<h_{t,j'}.
\label{eq:anchor-order}
\end{equation}
Records sharing an anchor are distinct but unordered; their row order is only a
stable storage tie-break, not evidence of a clause-level temporal order.
Figure~\ref{fig:anchor-schema} shows this distinction
between actions and the evidence that makes them interpretable.

\subsection{Identity: what is the same action?}

Each library entry is defined by four observable features:
\begin{equation}
\kappa(c)=\bigl(g(c),o^\star(c),\Delta s(c),p(c)\bigr),
\label{eq:identity}
\end{equation}
where $g$ is the proximal procedural goal, $o^\star$ the principal output,
$\Delta s$ the state transition, and $p$ the action-defining preconditions.
For an occurrence $c$ and entry $k$, the matching decision is therefore
conceptually
\begin{equation}
\operatorname{Match}(c,k)=\mathbf{1}\!\left[
g(c)\models g_k\ \land\ o^\star(c)\models o^\star_k\ \land\
\Delta s(c)\models\Delta s_k\ \land\ p(c)\models p_k\right].
\label{eq:match}
\end{equation}
Here $\models$ denotes an annotator's documented judgement under the frozen
entry definition; it is not an executable logical-entailment test.
Endpoint, identifier channel, record source, scenario, mutation object,
wording, and phase may describe an occurrence, but they do not decide its CPA
identity.  This is what allows one dictionary entry to remain meaningful across
surface interfaces.

Phase is retained only as an annotation scaffold.  It is assigned after the
CPA decision, has no role in Eq.~\ref{eq:identity}, and is excluded from the
archived observation sequence.  Similarly, a policy statement can ground an
action without becoming an action merely because it appears in a message.

\subsection{From open coding to a frozen instrument}

\begin{figure*}[t]
\centering
\resizebox{0.98\textwidth}{!}{
\begin{tikzpicture}[
  node distance=5mm and 6mm,
  box/.style={draw=CPAblue, rounded corners=2pt, fill=CPAlight,
    align=center, minimum height=15mm, text width=34mm, inner sep=3pt,
    font=\scriptsize},
  audit/.style={box, draw=CPAteal, fill=CPAteal!9},
  frozen/.style={box, draw=CPAorange, fill=CPAorange!10},
  sealed/.style={box, draw=CPAgray, dashed, fill=CPAgray!8, text=CPAgray},
  arrow/.style={-{Stealth[length=1.4mm]}, semithick, draw=CPAgray},
  note/.style={font=\scriptsize, text=CPAgray, align=center}
]
\node[audit] (r1) {\textbf{R1: open induction}\\[-1pt]
  32 hash-selected traces\\
  4 isolated contexts\\[-1pt]
  \tiny 973 records; 98 candidate strings};
\node[box, right=of r1] (review) {\textbf{Recorded consolidation}\\[-1pt]
  normalize candidate language\\
  merge / split / reject\\[-1pt]
  \tiny four-field definitions};
\node[audit, right=of review] (r2r5) {\textbf{R2--R5 APPLY-open}\\[-1pt]
  4 fresh 32-trace batches\\
  2 isolated contexts each\\[-1pt]
  \tiny proposals/abstentions reviewed};

\node[frozen, below=12mm of review] (v05) {\textbf{Frozen v0.5 candidate}\\[-1pt]
  library + prompt + schema + evaluator\\[-1pt]
  \tiny 24 canonical entries};
\node[audit, right=of v05] (r6) {\textbf{R6 fresh freeze audit}\\[-1pt]
  32 untouched traces; 990 records\\[-1pt]
  \tiny 0 new; 0 abstain; $F_{\rm mult}{=}.986$; $A{=}.982$};
\node[frozen, right=of r6] (v10) {\textbf{Freeze v1.0}\\[-1pt]
  immutable 24-entry library\\[-1pt]
  \tiny eligibility criteria all met};
\node[note, text width=32mm, left=of v05] (devnote)
 {library may change between R2 and R5: construction evidence, not IID evaluation};

\node[box, below=12mm of r6] (prod) {\textbf{Production APPLY}\\[-1pt]
  244 non-sealed trajectories\\[-1pt]
  \tiny 4,058 CPA records};
\node[box, right=of prod] (qa) {\textbf{QA and accounting}\\[-1pt]
  4,050 match; 5 abstain;\\ 3 diagnostics\\[-1pt]
  \tiny all schema/provenance checks pass};
\node[sealed, left=of prod] (sealed) {\textbf{Evaluation set}\\[-1pt]
  20 sealed trajectories\\[-1pt]
  \tiny unopened; excluded};
\node[note, text width=34mm, anchor=north] at ($(sealed.south)+(0,-2mm)$)
 {not sampled, annotated, or used to revise v1.0};

\draw[arrow] (r1) -- (review);
\draw[arrow] (review) -- (r2r5);
\draw[arrow] (r2r5.south) -- ++(0,-6mm) -| (v05.north);
\draw[arrow] (v05) -- (r6);
\draw[arrow] (r6) -- (v10);
\draw[arrow] (v10.south) -- ++(0,-6mm) -| (prod.north);
\draw[arrow] (prod) -- (qa);
\end{tikzpicture}}
\caption{\textbf{How the $\tau^3$-Retail library was built.} R1 is open
induction and R2--R5 are APPLY-open development rounds, separated by recorded
semantic decisions.  R6 evaluates the frozen candidate; the 20 sealed
evaluation traces were not opened or used to revise v1.0.}
\label{fig:pipeline}
\end{figure*}
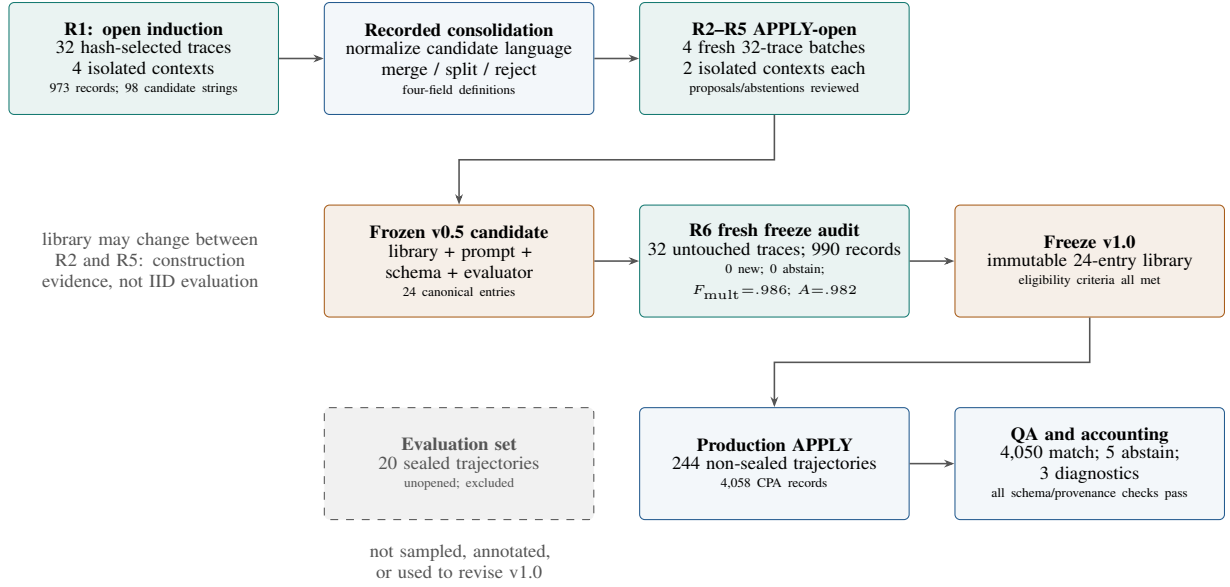

The construction process in Figure~\ref{fig:pipeline} follows two modes.  In
\induce{} mode, annotators work without a fixed library and either propose a
supported action name or abstain.  Candidate language is then normalized and
subjected to recorded merge, split, and reject decisions.  In \applymode{} mode,
the annotator works against a supplied library: it must use an exact entry when
the four-field definition fits, propose a new type only when no definition
fits, and abstain when the trace does not support a confident decision.  APPLY
results never alter a frozen library silently.

A freeze is a semantic decision supported by fresh application, not a count
threshold.  We freeze only when a fresh APPLY-open batch yields no surviving
new function after review, no recurring abstention or merge/split defect, and
no failure of the evidence or artifact checks.  Appendix~\ref{app:gates}
states the full criteria and the R6 evidence for each.

\subsection{Auditing the frozen instrument}

Before a corpus enters analysis, deterministic QA validates every referenced
ID, anchor, action-event list, context list, decision, label, and occurrence
identifier; it also checks anchor order, repeated-label preservation, and the
absence of sealed evaluation traces.  These checks establish that the data obey
the representation.  The separate audit asks whether isolated annotation runs
reproduce the representation on fresh trajectories.

For runs $r,s$, let $n_{th}^{(r)}$ be the number of
occurrences at trace--anchor pair $(t,h)$.  Anchored-occurrence multiplicity is
\begin{equation}
F_{\mathrm{mult}}=
\frac{2\sum_{t,h}\min\{n_{th}^{(r)},n_{th}^{(s)}\}}
     {\sum_{t,h}n_{th}^{(r)}+\sum_{t,h}n_{th}^{(s)}}.
\label{eq:multiplicity-f1}
\end{equation}
For label $k$, let $n_{thk}^{(r)}$ be its count at $(t,h)$ and define
\begin{align}
a_k&=\sum_{t,h}\min\{n_{thk}^{(r)},n_{thk}^{(s)}\},\\
p_k&=\sum_{t,h}\bigl(n_{thk}^{(r)}-n_{thk}^{(s)}\bigr)_+,
\quad q_k=\sum_{t,h}\bigl(n_{thk}^{(s)}-n_{thk}^{(r)}\bigr)_+,\\
A(k)&=\frac{2a_k}{2a_k+p_k+q_k},\qquad A_n(k)=a_k+p_k+q_k.
\label{eq:anchor-label-agreement}
\end{align}
Writing $a=\sum_k a_k$, $p=\sum_k p_k$, and $q=\sum_k q_k$, the reported
aggregate anchor--label agreement is
\begin{equation}
A=\frac{2a}{2a+p+q}.
\label{eq:aggregate-agreement}
\end{equation}
$F_{\mathrm{mult}}$ measures whether the runs place the same number of
occurrences at an anchor; $A(k)$ adds label identity.  Together, they capture
the two sources of disagreement introduced by same-message CPAs: how many
actions the runs find and which functions they assign.

These are symmetric overlap measures, not accuracy against a reference and
not chance-corrected reliability coefficients.  Shared omissions and shared
misinterpretations do not reduce them.  They also do not compare the semantic
sufficiency of cited evidence.  We report the unweighted mean of $A(k)$ only
over labels observed in at least one run, with an unobserved label shown as
undefined rather than as perfect agreement.

\section{Study design}
\label{sec:experiment}

The empirical study has three deliberately separate stages: constructing a
library, testing whether a frozen candidate can be reapplied on fresh data, and
using the frozen version to create a corpus.  Keeping these stages separate is
important.  Discovery rounds are allowed to change a definition; an evaluation
round is useful precisely because its library, prompt, schema, and batch have
already been fixed.

\subsection{Protocol summary}
\label{sec:protocol}

The operational sequence is: (1) select R1 and induce provisional actions in
isolated contexts; (2) consolidate candidates into a versioned library with
recorded decisions; (3) apply and revise through fresh R2--R5 batches;
(4) fix a freeze candidate, instructions, schema, evaluator specification, and
new batch before annotation; (5) run R6 and review the declared eligibility
criteria; and (6) freeze v1.0 and apply it to the remaining production traces,
retaining diagnostics.  Porting this procedure to another domain requires
new construction and validation; it does not imply portability of the retail
labels themselves.

\subsection{Data and sampling}

We work with the retail customer-service trajectories in $\tau^3$-Retail
\citep{yao2025taubench,sierra2026tau3}.  We retain the historical archive's
$\tau^3$-Retail designation; the exact upstream benchmark revision and the
configuration of the agent that generated these trajectories have not been
re-established from the analysis snapshot.  The results concern the archived
trajectories, not a newly run benchmark evaluation.  A trajectory contains customer turns, agent messages,
agent tool calls, and tool or environment results.  CPA annotates only the
agent's observable behaviour, while retaining the other events when they are
needed to establish an occurrence's interpretation.  Twenty benchmark
trajectories form a sealed evaluation set; they were neither opened nor used in
any library decision reported here.

The recorded split contains 456 trajectories: 20 reserved traces and 436
non-sealed traces.  A retrospective check of the actual ID lists verifies
that the six 32-trace construction/audit batches are pairwise disjoint
(192 traces), and that production contains the remaining 244.
This is trajectory separation, not task separation: the 32 R6 trajectories
cover 26 task IDs, of which 16 also occur in R1--R5 through different trials.

The first discovery sample (R1) consists of 32 traces selected by a recorded
hash procedure.  Four isolated annotation contexts received two disjoint
16-trace halves: A/B worked on one half and C/D on the other.  R1 produced 973
occurrence records: 965 provisional
proposals and eight abstentions.  Its 98 candidate label strings and 132
annotator-supplied definitions were inputs to consolidation, not an evaluation
score.  We normalized candidate language and recorded every merge, split, and
rejection before issuing the next library version.
The consolidation report proposed 29 canonical groups; the subsequent v0.1
library admitted 19 entries.  The historical R1 manifest lists a sampling
pool of 416; its relationship to the later 436-trace non-sealed partition
has not been re-established here.  We preserve the original field and base
the reported split accounting on the verified trace-ID lists.

\subsection{Development and the fresh audit}

R2--R5 are four further APPLY-open development rounds.  Each uses a fresh
32-trajectory batch and two isolated annotation contexts.  A proposed label,
abstention, or boundary observation in these rounds could trigger a documented
definition decision; these rounds therefore inform construction rather than
estimate final-system performance.  By the end of R5, review had produced a
24-entry candidate library, v0.5.  The complete development ledger, including
all proposal, abstention, and agreement counts, is provided in
Appendix~\ref{app:development}.
The applied versions were v0.1, v0.2, v0.3.1, and v0.4, respectively.

R6 is different.  Before annotation, we fixed the v0.5 library, the applied
prompt, occurrence schema, evaluator, and a 32-trajectory batch not used in development.
Two isolated contexts, C1 and C2, then applied that same package without
cross-visibility.  The audit manifest declares a Claude Code subagent,
\texttt{claude-opus-5}, and a one-million-token context window; these are
archived configuration declarations, not independently verified service
snapshots.  Temperature and top-$p$ were not pinned.  The freeze decision was made only after their outputs had
been parsed, compared, and checked against the criteria in
Appendix~\ref{app:gates}.  This design asks a focused question: given the
same frozen instrument, do fresh runs recover the same anchor-level action
multiset?

The historical input renderer retained assistant and customer text but clipped
tool arguments and tool-result strings to 110 characters.  Thus the audit
measures repeatability from those rendered views, not from complete tool
payloads.  Valid evidence references alone cannot establish that the visible
content was sufficient for an interpretation.

\subsection{Freeze and production application}

We promoted the candidate to v1.0 only if the R6 batch contained no surviving
new procedural function after review, no recurring unresolved abstention or
merge/split defect, and no action/context-reference, provenance,
artifact-integrity, or deterministic-QA failure.  These are semantic and
representational conditions, rather than a
universal threshold on a single agreement number.  The predeclared mechanical
checks validate parsing, vocabulary, event references, anchors, action/context
separation, multiplicity preservation, ordering, and sealed-set exclusion.

After the R6 decision, we applied v1.0 to 244 non-sealed trajectories.  The
production pass is an application of the frozen library: diagnostic proposals
and abstentions are retained as records for a later version, rather than
silently modifying v1.0.  The complete artifact map, reconstruction checks,
and annotation-configuration record are in Appendix~\ref{app:repro}.

\subsection{Retrospective validation analyses}
\label{sec:retrospective}

After the original freeze, we reparse the four raw R6 response files,
verify their equality to the archived 990 normalized records, and recompute
all agreement counts.  The additional analyses are exploratory; they were
not original freeze criteria.  We stratify by source anchor type, report all
24 labels with support, and construct a 95\% percentile interval by resampling
the 26 task clusters 5,000 times (seed 20260919), retaining both contexts and
all observed trials within each sampled task.  The interval describes
variation over this empirical task sample, not uncertainty over model runs
or human correctness.

We also merge the existing labels into eight explicit functional families,
and into a single action category, preserving every occurrence and anchor.
For a structural baseline, each native agent event after the initial greeting
contributes one occurrence.  Two simple labeling baselines map tool names to
CPA entries, with either one occurrence per call or consecutive same-endpoint
retrievals grouped at their first call; intervening customer and tool results
do not break a group, but an assistant message or different tool does.
No argument, result, or message content is interpreted.  These baselines are
compared separately to C1 and C2 as LLM outputs, not as gold labels.
Appendix~\ref{app:validation} gives the mappings, results, and outstanding
human and downstream evaluation designs.

We additionally compare exact context-event reference sets within each
trace--anchor--label key, preserving occurrence multiplicity.  Context order
is ignored; the cited IDs must otherwise agree.  A separate check also
requires the archived action-span endpoint to agree.  These are stricter
structural comparisons, not tests of whether different evidence selections
are semantically equivalent or sufficient.

\section{Results}
\label{sec:results}

The study produces both a repeatability result for the frozen annotation
protocol and a provenance-linked retail action corpus.  We report them in that
order: the first establishes how the instrument behaved on new data, and the
second describes what its fixed version yielded at scale.

\subsection{Paired outputs show high conditional overlap}

On R6, C1 returned 499 occurrences and C2 returned 491, for 990 records in
total.  Every record was a \match{} decision: neither isolated context
proposed a new CPA or abstained.  The two runs achieved anchored-occurrence
multiplicity $F_{\mathrm{mult}}=0.986$.  Of 354 anchors shared by the two
contexts, 340 carried exactly the same number of occurrences, and no anchor
appeared in only one context.  At the level of anchor--label multisets,
aggregate agreement was $A=0.982$ ($a=486$, $p=13$, $q=5$).

These figures are meaningful because the representation permits several
occurrences at one assistant-message anchor.  An ordinary one-label-per-turn
comparison would erase exactly the distinction CPA is intended to preserve.
Here, $F_{\mathrm{mult}}$ measures overlap in the number of anchored
occurrences, while $A$ asks whether the runs assigned the same
functions.  The R6 result therefore supports the claim that the fixed v0.5
instrument behaved consistently on its fresh batch.  The full R2--R6
development history and R6 comparison ledger are reported in
Appendix~\ref{app:development}.

\subsection{Aggregate agreement hides uneven support and local disagreement}

The retrospective task-cluster interval for aggregate $A$ is
$[\CPAAuditLower,\CPAAuditUpper]$.  The macro-average over the 23 observed
labels is \CPAMacroAgreement; the remaining label,
\path{SEQUENCE_MUTATION_STEPS}, is unobserved and has no estimable R6
agreement.  Three other labels have only one occurrence per context, and
\path{DECLARE_INFORMATION_UNAVAILABLE} has two.  Perfect overlap for
these entries is weak evidence about their stability.

The weakest observed entry is \path{BIND_PARAMETER_FROM_RECORD}:
C1 emits 27 occurrences and C2 emits 22, with 20 matched, giving
$A(k)=40/49=0.816$.  Its nine unmatched occurrences account for half of all
18 unmatched records.  \path{DECLINE_OUT_OF_SCOPE_REQUEST} and
\path{SURVEY_ORDER_PORTFOLIO} score 0.833 and 0.889, respectively.
All disagreements occur at assistant-message anchors: the 185 tool-anchored
occurrences in each context agree exactly, whereas the 314/306 message
occurrences yield $A=\CPAMessageAgreement$.  Anchor modality is observable;
it is not itself a validated measure of semantic difficulty.
Appendix~\ref{app:validation} gives all label counts and the residual
multisets at the \CPADisagreementAnchors{} disputed anchors.

\subsection{Label overlap does not ensure identical evidence references}

Requiring the same context-event ID set as well as the same trace, anchor,
and label reduces the matched count from 486 to \CPAContextMatched{}.
The resulting symmetric overlap is
$A_{\rm ctx}=2\times395/(499+491)=\CPAContextAgreement$.
Requiring the archived action-span endpoint alone leaves the original 486
matches unchanged; requiring both span endpoint and context set yields
395 matches.  Thus the headline $A=0.982$ leaves substantial variation in
context-reference selection unmeasured.

Different reference sets can be equally sufficient, while identical sets
can both be insufficient.  This result is neither a semantic error rate nor
a complete-record accuracy score: confidence, phase, outcomes, and other
metadata are not part of this comparison.  It identifies evidence selection
as a distinct validation target, especially given the clipped tool payloads.

\subsection{Coarsening and endpoint rules delimit the current evidence}

Mapping the 24 labels into eight functional families raises $A$ from 0.982
to \CPACoarseAgreement.  Collapsing every label to one category produces the
same 0.986 score, despite eliminating every label distinction.  This is a
controlled remapping of existing outputs, not a new annotation experiment.
It demonstrates why agreement cannot select the vocabulary's granularity.
Both R6 contexts also reported that \path{SOLICIT_MUTATION_INPUT}
covered distinct-looking value requests, choices, ratifications, and execution
authorizations, despite agreeing on all 72 occurrences per context.
A finer semantic taxonomy has not yet been independently annotated or tested
for utility.

The grouped endpoint-rule baseline matches all 185 tool-anchored occurrences
in each context and emits one additional occurrence; its overlap restricted
to tool anchors is 0.997.  Against each complete annotation, those same
185 matches cover only 37.1\%/37.7\% of C1/C2 occurrences.  Thus much of the
tool portion is cheaply recoverable, while the additional CPA labels describe
work in assistant messages.  Native-event multiplicity and full baseline
results appear in Table~\ref{tab:baselines}.  These are reconstruction
comparisons to LLM annotations; lower baseline overlap does not demonstrate
that CPA improves failure diagnosis or any other downstream task.

\subsection{A reproducible query over co-anchored message actions}
\label{sec:query-example}

To make the representation's use concrete, we query the archived R6 outputs
for anchors carrying both \path{PRESENT_VARIANT_OPTIONS} and
\path{SOLICIT_MUTATION_INPUT}.  Each context returns the same
\CPAQueryCount{} anchors in \CPAQueryTraceCount{} trajectories.  This query
needs two functions at a shared anchor; a single endpoint label does not
encode that conjunction.  The query is a descriptive demonstration on LLM
outputs, not an evaluated improvement over a content-based rule or raw-trace
search.

For example, at \texttt{T103-3:36}, the assistant reports an address update,
lists three red luggage variants, and asks the customer to choose a variant
and a payment method.  Both contexts record an acknowledgement, variant
presentation, and mutation-input solicitation at that anchor.  However, they
cite different context events for the solicitation: C1 cites events 7 and 31,
whereas C2 cites 31 and 35.  Their anchor--label agreement for this message is
perfect despite this evidence-reference difference.  Checking whether the
underlying claims and cited evidence are sufficient still requires inspection
of the source payloads.

Across all R6 outputs, C1 and C2 assign multiple occurrences to
\CPAMultiAnchorCOne{} and \CPAMultiAnchorCTwo{} assistant-message anchors,
respectively, out of 354 occupied anchors per context.  These are counts of
annotated multiplicity, not independently verified counts of real actions.
The query script emits the matching trace/event IDs and checks the equality
of the returned anchor sets.

\subsection{Frozen application yields 4,058 archived records}

The v1.0 application completed all 244 expected non-sealed trajectories and
created 4,058 CPA records.  Its decision accounting was
\begin{equation}
\widehat p_{\mathrm{match}}=\frac{4050}{4058}=0.9980,
\qquad
\widehat p_{\mathrm{diagnostic}}=
\frac{5+3}{4058}=0.0020,
\label{eq:production-rates}
\end{equation}
where the diagnostic outcomes comprise five \abstain{} records and three
\propose{} records.  The three proposals were differently worded
out-of-domain question types.  They remain visible diagnostics for a later
version, preserving the meaning of a frozen v1.0 corpus.

Every one of the 24 dictionary entries occurs in production.  A trajectory
contains 16.63 records on average (range 6--33), and repeated labels are
preserved as distinct occurrences rather than deduplicated.  All output records
parsed; all referenced evidence IDs were valid; empty context lists were
accepted where the schema allowed them; and all remaining deterministic
provenance and ordering checks passed.  The detailed accounting and QA
interpretation are in Appendix~\ref{app:development}.

\subsection{CPA exposes a broad, uneven repertoire of retail procedures}

Figure~\ref{fig:frequency} makes the resulting action layer tangible.  The
highest-frequency functions trace a recognizable retail-service loop: obtain
identity credentials, resolve a customer and request referent, retrieve records,
obtain remaining mutation input, execute an authorized change, and acknowledge
the result.  At the same time, the corpus retains less common but meaningful
branches, including sequencing required changes (12 occurrences) and declaring
that information is unavailable (13 occurrences).  Those cases would be easy
to lose in a representation based only on endpoint names or final task status.

The distribution is descriptive rather than a leaderboard of labels.  Its
value is that analysts can now ask distributional or process questions over
procedural functions while returning to the source event IDs behind any count.
The operational definitions and exact frequency of every entry appear in
Appendix~\ref{app:dictionary}.

\begin{figure*}[!htbp]
\centering
\resizebox{0.90\textwidth}{!}{
\begin{tikzpicture}[font=\scriptsize]
\newcommand{\freqrow}[3]{%
  \node[anchor=east,font=\ttfamily\scriptsize] at (63mm,#1mm) {#2};%
  \pgfmathsetmacro{\barwidth}{#3*0.14}%
  \pgfmathsetmacro{\yb}{#1-1.35}%
  \pgfmathsetmacro{\xcount}{66+\barwidth}%
  \fill[CPAblue] (65mm,\yb mm) rectangle ++(\barwidth mm,2.7mm);%
  \node[anchor=west,font=\scriptsize] at (\xcount mm,#1mm) {#3};%
}
\node[anchor=west,font=\bfseries] at (0,119mm)
  {Production occurrences per frozen CPA};
\node[anchor=west,font=\scriptsize,text=CPAgray] at (0,115mm)
  {Each bar is a count of occurrence records; labels are sorted by frequency.};
\draw[->, CPAgray] (65mm,3mm) -- (138mm,3mm);
\node[anchor=north,font=\tiny,text=CPAgray] at (101.5mm,-1mm) {occurrence records};
\foreach \x/\lab in {0/0,14/100,28/200,42/300,56/400,70/500}
 {\pgfmathsetmacro{\xtick}{65+\x}
  \draw[CPAgray!40] (\xtick mm,3mm) -- (\xtick mm,114mm);
  \node[anchor=north,font=\tiny] at (\xtick mm,2mm) {\lab};}
\freqrow{111}{SOLICIT\_MUTATION\_INPUT}{499}
\freqrow{106.5}{REQUEST\_IDENTITY\_CREDENTIAL}{399}
\freqrow{102}{EXECUTE\_AUTHORIZED\_MUTATION}{368}
\freqrow{97.5}{RETRIEVE\_ORDER\_RECORD}{293}
\freqrow{93}{RESOLVE\_REQUEST\_REFERENT}{285}
\freqrow{88.5}{RESOLVE\_CUSTOMER\_IDENTITY}{268}
\freqrow{84}{ACKNOWLEDGE\_MUTATION\_COMMIT}{267}
\freqrow{79.5}{COMPUTE\_SETTLEMENT\_AMOUNT}{248}
\freqrow{75}{RETRIEVE\_ACCOUNT\_PROFILE}{241}
\freqrow{70.5}{RETRIEVE\_PRODUCT\_VARIANT\_SET}{184}
\freqrow{66}{BIND\_PARAMETER\_FROM\_RECORD}{175}
\freqrow{61.5}{SCREEN\_ROUTE\_ADMISSIBILITY}{152}
\freqrow{57}{ANSWER\_CUSTOMER\_QUERY}{129}
\freqrow{52.5}{SELECT\_VARIANT\_MEETING\_CONSTRAINTS}{119}
\freqrow{48}{PRESENT\_VARIANT\_OPTIONS}{97}
\freqrow{43.5}{SCREEN\_CANDIDATE\_SATISFACTION}{81}
\freqrow{39}{SURVEY\_ORDER\_PORTFOLIO}{75}
\freqrow{34.5}{OFFER\_ALTERNATIVE\_COURSES}{60}
\freqrow{30}{DECLINE\_OUT\_OF\_SCOPE\_REQUEST}{38}
\freqrow{25.5}{RETRIEVE\_PRODUCT\_CATALOG\_INDEX}{16}
\freqrow{21}{TRANSFER\_TO\_HUMAN\_AGENT}{16}
\freqrow{16.5}{ELICIT\_GOAL\_SELECTION}{15}
\freqrow{12}{DECLARE\_INFORMATION\_UNAVAILABLE}{13}
\freqrow{7.5}{SEQUENCE\_MUTATION\_STEPS}{12}
\node[anchor=west,font=\tiny,text=CPAgray,align=left] at (0,-6mm)
 {Total: 4,050 \match{} records across 244 trajectories.  Five abstentions and
  three proposed-label diagnostics are excluded; counts are descriptive.};
\end{tikzpicture}}
\caption{\textbf{Production frequency of every frozen CPA.} Bars sum to the
4,050 \match{} records across 244 $\tau^3$-Retail trajectories; five
abstentions and three proposed-label diagnostics are excluded.  The figure
reveals the procedural repertoire exercised by the corpus; it is not a
label-quality ranking.}
\label{fig:frequency}
\end{figure*}

\section{Discussion}
\label{sec:discussion}

\subsection{An action layer between traces and behavioural analysis}

Raw trajectories preserve fine-grained evidence but leave the procedural role
of an event implicit.  CPA supplies a middle layer: it turns observable
functions into queryable occurrences without discarding the messages, tool
calls, and results that support them.  That combination is intended to support
three kinds of future analysis.  It permits comparisons across interfaces that
serve the same procedural function, preserves repeated actions that a
set-valued summary would lose, and exposes an anchor-respecting partial order
for flow or process analyses.  In each case, an analyst can return from a
canonical label to its source evidence.

The design also makes a practical modelling choice explicit.  When a customer
response can serve as a value, a choice, a ratification, or authorization for a
pending change, the trace often lacks an observable state that separates those
latent distinctions.  The v1 library therefore uses
\path{SOLICIT_MUTATION_INPUT} as an observation-level merge.  This is a
deliberate boundary of the instrument, recorded in the dictionary and tested as
a dedicated R6 watch item, rather than a claim that those underlying activities
are psychologically identical.  Consequently, this label cannot by itself establish that explicit authorization
was obtained.  Likewise, a record named
\path{EXECUTE_AUTHORIZED_MUTATION} is an annotation decision, not a
verified policy-compliance judgement.  Similar choices can be revised transparently
in a later version because the current occurrences retain their provenance.

\subsection{Scope and next validation}

The empirical result is conditional repeatability of an annotation protocol for the observed
$\tau^3$-Retail trajectories.  R6 compares two isolated LLM contexts, not a
human-reference study, so its agreement measures establish protocol
consistency rather than semantic ground truth.  Shared omissions, common
mislabeling, and insufficient evidence can all survive perfect agreement.
The context-reference overlap of $0.798$, low per-label scores, and sparse
support reported here make those risks
specific.  The 110-character tool-payload rendering limit further prevents
equating valid provenance with complete evidence.

Production used one annotation stream with a documented model-configuration
change: s03--s04 used the earlier configuration, and the other shards used
the later one.  Temperature and top-$p$ were not pinned in either production
or R6; a declared model-family string is not a verified deployment snapshot.
The 24-entry library was
constructed for this retail environment.  The three production proposal
diagnostics and five abstentions illustrate why its boundaries should remain
visible when the corpus is reused.

The outstanding validation has four parts.  First, two humans should annotate
complete traces independently before seeing model outputs, with a third
adjudicator separating missed actions, wrong labels, multiplicity errors, and
insufficient evidence.  Second, an independently applied finer taxonomy must
test whether splitting mutation-input requests improves distinctions relevant
to a task, rather than merely changing agreement.  Third, a blinded failure
diagnosis study must compare native traces, deterministic labels, and CPA
under the same information and time budgets.  Fourth, repeated runs with
recorded configurations and a task-disjoint second domain must distinguish
within-model repeatability from between-model robustness and transfer.
Appendix~\ref{app:validation} specifies these proposed studies; none is
reported as a completed result.  The historical 20-trace reserve was excluded
from this study; a future evaluation must verify its then-current exposure
status before calling it untouched.  Any sequence model built on
CPA should respect Eq.~\ref{eq:anchor-order}: same-message occurrences are
co-anchored observations, not evidence of a total within-message order.

\section{Conclusion}

Canonical Procedural Actions turn $\tau^3$-Retail traces into an auditable
action layer whose labels remain connected to their event-level evidence.  The
paper contributes the representation, construction and freeze protocol,
conditional repeatability evaluation, applied prompt summary, criteria, and a 24-entry
dictionary applied to 4,058 occurrence records.  It offers a concrete foundation for
studying how tool-use agents act---with the evidence needed to question,
reproduce, or refine that measurement in future work.

\section*{AI-use statement}
LLM contexts generated the corpus annotations and audits described in this
paper; these are not human annotations.  Generative AI also assisted manuscript
drafting, editing, and retrospective analysis code.  No AI output is counted
as an independent human validation result.

\FloatBarrier
\bibliographystyle{plainnat}
\bibliography{references}

\clearpage
\appendix
\section{Applied annotation prompt and record format}
\label{app:prompt}

This appendix gives a faithful, formatting-normalized summary of the
instructions applied in the final $\tau^3$-Retail process.  It is included so
that the operational definition is inspectable in the paper.  The accompanying
source bundle contains the frozen codebook, archived schema, and freeze
manifest (Appendix~\ref{app:repro}).  The text below is an \emph{application protocol},
not a claim that a prompt alone validates labels.

\subsection{Core instruction}

\begin{lstlisting}
You are annotating retail customer-service agent trajectories.

Segment agent behaviour into Canonical Procedural Actions (CPAs).
The library is FROZEN: do not edit entries, rename them, or stretch a
definition merely to make an event fit.

A CPA is a goal-directed action above an individual operation and below the
whole task.  Decide CPA identity only from:
  (1) proximal procedural goal
  (2) principal output
  (3) procedural state transition
  (4) action-defining preconditions

These NEVER make different CPAs: tool endpoint, identifier channel, record
source, scenario, mutation object, wording, or phase.

Record only AGENT behaviour. Customer turns and tool results may be context
evidence, but are never action events. The greeting is not an action.

ACTION_ANCHOR: first AGENT event realizing the CPA.
ACTION_EVENT_IDS: all AGENT events jointly realizing it.
CONTEXT_EVENT_IDS: customer/tool/environment events establishing the goal,
input, output, state change, or outcome. Empty context is permitted.

One agent message frequently realizes several CPAs. Record one occurrence per
realized function. Do not collapse them; do not invent them. A clause is a
separate occurrence only when it has its own principal output or state change.

For every occurrence choose exactly one:
  MATCH_EXISTING: a supplied library definition fits; use its exact label.
  PROPOSE_NEW: the occurrence is a genuine action and no definition fits.
  ABSTAIN: the occurrence/action type cannot be determined confidently.

Match definitions, never label wording. Do not propose a new CPA because of a
different tool, scenario, mutation type, input channel, or wording.
Do not consult another annotator's output. Do not modify the library.
\end{lstlisting}

The prompt places phase after CPA selection.  It permits one of
\textsc{Plan}, \textsc{Retrieve}, \textsc{Inspect}, \textsc{Extract},
\textsc{Verify}, \textsc{Write}, \textsc{Synthesize}, \textsc{Repair}, or
\textsc{Handoff}, but explicitly says that phase must not influence the
identity decision.  Final production sequences do not contain phase.
The applied prompt does not list an \textsc{Unresolved} phase; we do not add
one retroactively.  Its three confidence fields refer to action boundaries,
the auxiliary phase, and CPA type, respectively, on a 0.00--1.00 scale with
two decimals.  These are annotator self-reports, not calibrated probabilities
or additional quality measurements.

\subsection{Required response layout}

The source instruction asked for one pipe-delimited row per occurrence in trace
and anchor order, followed by definitions only for proposed labels and a short
library-observation section.  The core row was:

\begin{lstlisting}
TRACE | ANCHOR-LASTACTIONEVENT | CONTEXT_IDS | LABEL | PHASE | DECISION |
OUTCOME | BOUNDARY_CONF | PHASE_CONF | TYPE_CONF | REVIEW | REVIEW_REASON
\end{lstlisting}

Rows that share an anchor use a stable storage tie-break only; their relative
row order carries no within-message temporal interpretation.

For a \match{} row, \path{LABEL} must be an exact frozen-library label.  For
a \propose{} row, it is an uppercase-snake-case candidate; the required
definition supplies goal, inputs, output, state transition, preconditions,
inclusion/exclusion criteria, positive occurrences, nearest library entry, and
the four-field reason it does not match.  \texttt{SECTION 3} allows only
\path{TOO_BROAD} and \path{SHOULD_MERGE} observations (or
\path{NONE}); it is a route for codebook review, not an edit channel.

The raw row's first--last interval is normalized into the frozen v1 occurrence
schema.  Its first agent event becomes \textsc{ActionAnchor}; all ordered agent
events in the interval become \textsc{ActionEventIds}; and interleaved
customer/tool/environment events become \textsc{ContextEventIds}.  This
normalization is why a tool result can support an action without becoming part
of its action-event list.  It preserves a same-message multi-CPA occurrence as
separate records instead of turning it into a single span label.

\subsection{Worked schematic example}

\begin{lstlisting}
Customer: "Can I change this order to blue?"
Agent tool: retrieve order record
Tool result: order and available variants
Agent: "Blue is available. Would you like me to make that change?"

RETRIEVE_ORDER_RECORD        anchor=tool call, context=tool result
PRESENT_VARIANT_OPTIONS      anchor=agent message, context=tool result
SOLICIT_MUTATION_INPUT       anchor=same agent message, context=customer turn
\end{lstlisting}

The final two rows share an anchor but have different outputs: exposing the
option space versus obtaining the customer's authorization/input.  Their order
within the message is not recorded as an observation.  This is schematic only;
real annotations must use source event IDs and the exact frozen library
definition.

\section{Actual QA, promotion, and freeze criteria}
\label{app:gates}

This appendix reports the criteria actually used for the frozen v1.0 corpus.
They should not be confused with a generic or preregistered accuracy threshold.
In particular, support counts and $A(k)$ are descriptive evidence; neither can
by itself admit a label or establish semantic truth.

\subsection{Hard per-output mechanical checks}

Every production record must satisfy the record contract before it enters the
corpus.  Table~\ref{tab:mechanical-gates} lists the deterministic checks.  A
failed record is not silently repaired by changing its semantic label; it must
be rejected, re-annotated, or recorded as an explicit diagnostic according to
the versioned process.

\begin{longtable}{p{0.29\linewidth}p{0.36\linewidth}p{0.24\linewidth}}
\caption{Mechanical QA criteria for a frozen CPA output.}
\label{tab:mechanical-gates}\\
\toprule
Check & Deterministic condition & Purpose \\
\midrule
\endfirsthead
\toprule
Check & Deterministic condition & Purpose \\
\midrule
\endhead
Parse and vocabulary & Every output parses; decision and phase values belong
to their controlled vocabularies; each \match{} label is in v1.0. & Prevent
unreadable or out-of-library records. \\
Occurrence identity & Occurrence IDs are unique; repeated labels remain
separate occurrences; anchors are nondecreasing in trace order. & Preserve
multiset identity rather than collapse repetitions. \\
Action provenance & \textsc{ActionAnchor} exists, is an agent event, and is
the first \textsc{ActionEventId}; all action IDs are agent-only and ordered. &
Enforce the anchor-aware action contract. \\
Context provenance & Every cited context ID exists in the trace; an empty
context is accepted when valid. & Keep observable evidence inspectable without
requiring an artificial context event. \\
Schema separation & Action and context fields are present under the frozen
schema; no phase is emitted into the archived observation sequence. & Prevent
unlicensed semantic fields from entering downstream data. \\
Evaluation separation & No sealed evaluation trace is present in production.
& Preserve the declared holdout boundary. \\
\bottomrule
\end{longtable}

All of these checks pass for the 4,058 production records.  This outcome means
the records conform to the specified representation.  It does not mean that a
model's proposed label is a human-validated truth.

\subsection{Candidate and definition decisions}

An APPLY-open candidate is actionable only when no current library definition
fits its four identity fields.  Candidate spelling is normalized, but string
similarity is never an automatic semantic decision.  A recorded review can
merge synonyms, split an over-broad definition, reject an unsupported
candidate, or retain an observation-level merge when the discriminating state
is not observable.  A candidate found after a freeze remains a diagnostic for a
subsequent version; it cannot alter the frozen corpus retroactively.

The freeze process rejects a proposed universal admission equation such as
``occurrence count $\geq N$ and trajectory count $\geq R$.''  Counts across
independent samples and contexts are retained as support information, including
for rare labels, but definition stability and reviewed semantic distinctness are
the relevant conditions.  Likewise, $F_{\rm mult}$ and $A(k)$ in
Section~\ref{sec:method} diagnose repeatability at an anchor; they do not
replace merge/split review.

\subsection{v1 freeze decision}

Table~\ref{tab:freeze-gates} gives the actual v0.5-to-v1 eligibility decision
on the R6 fresh 32-trace audit.  The audit was run with frozen library, prompt,
schema, evaluator, and batch artifacts.  The final v1.0 step is a versioned
freeze of that eligible library, not a new annotation experiment.

\begin{longtable}{p{0.31\linewidth}p{0.18\linewidth}p{0.40\linewidth}}
\caption{Freeze criteria and evidence from the final audit.}
\label{tab:freeze-gates}\\
\toprule
Criterion & Verdict & Evidence \\
\midrule
\endfirsthead
\toprule
Criterion & Verdict & Evidence \\
\midrule
\endhead
No recurrent surviving \propose{} function & Pass & Zero proposals across 990
occurrences, following a prior round with zero proposals; therefore none
survives semantic review. \\
No repeated \abstain{} / ambiguity pattern & Pass & Zero abstentions in the
entire R6 batch. \\
No reproducible merge/split defect & Pass & No defect reproduced by both
isolated contexts under the documented defect rule.  A single-context
\path{DECLINE_OUT_OF_SCOPE_REQUEST} versus
\path{SCREEN_ROUTE_ADMISSIBILITY} observation remains a watch item. \\
No definition-level contradiction & Pass & No invalid label, phase, schema, or
reference; reported breadth was licensed by the relevant definitions. \\
Stable observation-level merge & Pass &
\path{SOLICIT_MUTATION_INPUT}: 144 occurrences across the paired outputs
over all 32 traces,
$A(k)=1.000$, identical 70 shared anchors with exact count agreement. \\
Mechanical evidence coverage & Pass & Every record cites context and every
cited ID exists.  This does not test semantic sufficiency or completeness of
the rendered tool payload. \\
Frozen artifacts unchanged & Pass & Library, prompt, schema, evaluator, and
batch manifest hashes match the pre-audit record. \\
\bottomrule
\end{longtable}

The resulting audit metrics---$F_{\rm mult}=0.986$ and aggregate
$A=0.982$---are reported because they make the degree of contextual
reproducibility visible.  They are not threshold values that can turn a
semantically defective codebook into a valid one.

The historical pass for the mutation-input merge is conditional on the
supplied definition.  Both contexts reported its breadth in their library
observations.  Those observations remain concerns about granularity even
though no change was required by the historical freeze rule.

\section{Development and QA detail}
\label{app:development}

This appendix retains the construction ledger and mechanical accounting behind
the main-paper results.  R2--R5 are development rounds: library versions could
change after recorded review, so their values describe the path to a stable
instrument rather than a performance trajectory.  R6 is the only fresh,
fixed-instrument evaluation in this table.

\subsection{Round-by-round construction record}

\begingroup
\small
\setlength{\tabcolsep}{2pt}
\begin{longtable}{p{0.09\linewidth}p{0.18\linewidth}p{0.14\linewidth}p{0.11\linewidth}p{0.20\linewidth}p{0.08\linewidth}p{0.07\linewidth}}
\caption{Discovery, development, and fresh-audit record.  Proposal counts are
occurrences; parenthetical values give the number of distinct proposed labels
when applicable.}\label{tab:development-ledger}\\
\toprule
Round & Role & Traces / contexts & Records & \propose{} / \abstain{} &
$F_{\rm mult}$ & $A$ \\
\midrule
\endfirsthead
\toprule
Round & Role & Traces / contexts & Records & \propose{} / \abstain{} &
$F_{\rm mult}$ & $A$ \\
\midrule
\endhead
R1 & Open induction & 32 / 4 isolated & 973 & 965 / 8 (98 candidate strings) &
-- & -- \\
R2 & APPLY-open development & 32 / 2 isolated & 992 & 24 / 0 (8 labels) &
.964 & .950 \\
R3 & APPLY-open development & 32 / 2 isolated & 1,125 & 7 / 3 (4 labels) &
.964 & .946 \\
R4 & APPLY-open development & 32 / 2 isolated & 1,039 & 5 / 2 (3 labels) &
.964 & .949 \\
R5 & APPLY-open development & 32 / 2 isolated & 1,076 & 0 / 1 &
.974 & .965 \\
R6 & Fresh v0.5 audit & 32 / 2 isolated & 990 & 0 / 0 &
.986 & .982 \\
\bottomrule
\end{longtable}
\endgroup

R1's 98 candidate strings came with 132 annotator-supplied label definitions.
The consolidation record normalized these proposals and recorded their merge,
split, retention, or rejection before any subsequent library version was
issued.  The 24-entry v0.5 candidate used in R6 was already fixed before the
new batch was opened.  Thus the improvement visible in the ledger should be
read as evidence of construction and review, whereas R6 tests whether the
resulting instrument reproduces on fresh trajectories.

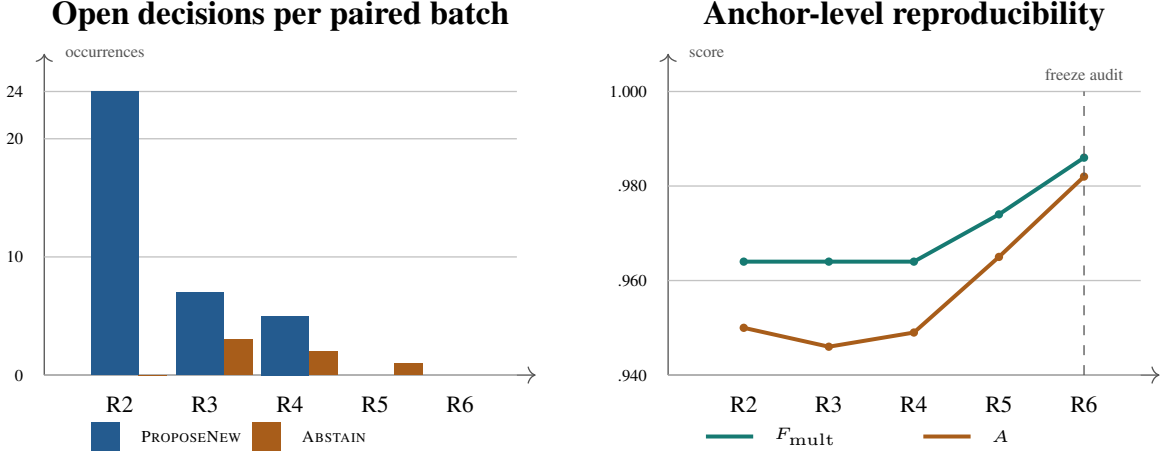
\begin{figure*}[t]
\centering
\resizebox{0.94\textwidth}{!}{
\begin{tikzpicture}[font=\scriptsize]
\node[font=\bfseries] at (25mm,38mm) {Open decisions per paired batch};
\draw[->, CPAgray] (0,0) -- (0,34mm);
\draw[->, CPAgray] (0,0) -- (52mm,0);
\foreach \y/\lab in {0/0,12.5/10,25/20,30/24}
  {\draw[CPAgray!35] (0,\y mm) -- (50mm,\y mm);
   \node[anchor=east,font=\tiny] at (-1mm,\y mm) {\lab};}
\foreach \x/\lab in {8/R2,17/R3,26/R4,35/R5,44/R6}
  {\node[anchor=north] at (\x mm,-1mm) {\lab};}
\fill[CPAblue] (5mm,0) rectangle (10mm,30mm);
\fill[CPAblue] (14mm,0) rectangle (19mm,8.75mm);
\fill[CPAblue] (23mm,0) rectangle (28mm,6.25mm);
\fill[CPAorange] (10mm,0) rectangle (13mm,0mm);
\fill[CPAorange] (19mm,0) rectangle (22mm,3.75mm);
\fill[CPAorange] (28mm,0) rectangle (31mm,2.5mm);
\fill[CPAorange] (37mm,0) rectangle (40mm,1.25mm);
\node[anchor=west,font=\tiny,text=CPAgray] at (1mm,34mm) {occurrences};
\fill[CPAblue] (5mm,-8mm) rectangle (8mm,-5mm);
\node[anchor=west,font=\tiny] at (9mm,-6.5mm) {\propose{}};
\fill[CPAorange] (22mm,-8mm) rectangle (25mm,-5mm);
\node[anchor=west,font=\tiny] at (26mm,-6.5mm) {\abstain{}};

\begin{scope}[xshift=66mm]
\node[font=\bfseries] at (25mm,38mm) {Anchor-level reproducibility};
\draw[->, CPAgray] (0,0) -- (0,34mm);
\draw[->, CPAgray] (0,0) -- (52mm,0);
\foreach \y/\lab in {0/.940,10/.960,20/.980,30/1.000}
  {\draw[CPAgray!35] (0,\y mm) -- (50mm,\y mm);
   \node[anchor=east,font=\tiny] at (-1mm,\y mm) {\lab};}
\foreach \x/\lab in {8/R2,17/R3,26/R4,35/R5,44/R6}
  {\node[anchor=north] at (\x mm,-1mm) {\lab};}
\draw[CPAteal, very thick] (8mm,12mm) -- (17mm,12mm) -- (26mm,12mm) --
  (35mm,17mm) -- (44mm,23mm);
\foreach \x/\y in {8/12,17/12,26/12,35/17,44/23}
  {\fill[CPAteal] (\x mm,\y mm) circle (1.25pt);}
\draw[CPAorange, very thick] (8mm,5mm) -- (17mm,3mm) -- (26mm,4.5mm) --
  (35mm,12.5mm) -- (44mm,21mm);
\foreach \x/\y in {8/5,17/3,26/4.5,35/12.5,44/21}
  {\fill[CPAorange] (\x mm,\y mm) circle (1.25pt);}
\node[anchor=west,font=\tiny,text=CPAgray] at (1mm,34mm) {score};
\draw[CPAteal, very thick] (4mm,-6.5mm) -- (9mm,-6.5mm);
\node[anchor=west,font=\tiny] at (10mm,-6.5mm) {$F_{\rm mult}$};
\draw[CPAorange, very thick] (27mm,-6.5mm) -- (32mm,-6.5mm);
\node[anchor=west,font=\tiny] at (33mm,-6.5mm) {$A$};
\draw[CPAgray, dashed] (44mm,0) -- (44mm,30mm);
\node[anchor=south,font=\tiny,text=CPAgray] at (44mm,30mm) {freeze audit};
\end{scope}
\end{tikzpicture}}
\caption{\textbf{Development record and fresh-audit result.} Each APPLY-open
round uses a fresh 32-trace batch, but the library changes between R2 and R5.
The R6 point is the fixed-library audit: zero proposals, zero abstentions,
$F_{\mathrm{mult}}=.986$, and $A=.982$.}
\label{fig:validation-ledger}
\end{figure*}

\subsection{R6 comparison and freeze evidence}

The R6 outputs used the same fixed library, prompt, schema, evaluator, and
batch manifest.  C1 produced 499 records and C2 produced 491.  Every row
parsed, all decisions were \match{}, all label and phase values were valid,
and all referenced evidence IDs existed in the corresponding trace.  Table
\ref{tab:r6-detail} exposes the anchor-level quantities behind the two main
agreement statistics.  The full semantic and mechanical freeze criteria,
including the retained watch item, are in Appendix~\ref{app:gates}.

\begin{table}[t]
\centering
\small
\caption{Detailed evidence from the fixed R6 audit.}
\label{tab:r6-detail}
\begin{tabularx}{\linewidth}{p{0.45\linewidth}Y}
\toprule
Quantity & Value \\
\midrule
Frozen inputs & v0.5 library, prompt, schema, evaluator, and batch manifest \\
C1 / C2 output records & 499 / 491 \\
Decision outcomes & 990 \match{}; 0 \propose{}; 0 \abstain{} \\
Shared anchors / unique-to-one-context anchors & 354 / 0 \\
Shared anchors with exact occurrence count & 340 \\
Matched anchored multiplicities & 488; $F_{\mathrm{mult}}=.986$ \\
Anchor--label multiset counts & $a=486$, $p=13$, $q=5$; $A=.982$ \\
Evidence and artifact checks & Valid-ID and evidence-completeness rates 1.0;
frozen artifact hashes match \\
\bottomrule
\end{tabularx}
\end{table}

\subsection{Production accounting and deterministic QA}

\begin{figure}[t]
\centering
\resizebox{0.97\linewidth}{!}{
\begin{tikzpicture}[
  node distance=7mm and 10mm,
  box/.style={draw=CPAblue, rounded corners=2pt, fill=CPAlight,
    align=center, minimum height=16mm, text width=32mm, inner sep=3pt,
    font=\scriptsize},
  matchbox/.style={box, draw=CPAteal, fill=CPAteal!10},
  diagbox/.style={box, draw=CPAorange, fill=CPAorange!10},
  sealed/.style={box, draw=CPAgray, dashed, fill=CPAgray!8, text=CPAgray,
    text width=36mm},
  arrow/.style={-{Stealth[length=1.4mm]}, semithick, draw=CPAgray},
  holdout/.style={thin, dashed, draw=CPAgray}
]
\node[box] (traces) {\textbf{244 / 244}\\
  expected non-sealed\\ trajectories completed};
\node[box, right=of traces] (occ) {\textbf{4,058}\\
  occurrence records\\
  \tiny mean 16.63 / trajectory};
\node[matchbox, right=of occ, yshift=11mm] (match) {\textbf{4,050 MATCH}\\
  \textbf{99.80\%}\\
  \tiny exact v1.0 library label};
\node[diagbox, right=of occ, yshift=-12mm] (diag) {\textbf{8 diagnostics}\\
  5 \abstain{} + 3 \propose{}\\
  \tiny library remains unchanged};
\node[box, right=of match, yshift=-11mm] (qa) {\textbf{24 / 24 labels}\\
  exercised; all QA\\ checks pass};
\node[sealed, below=9mm of traces] (sealed)
  {\textbf{20 sealed evaluation traces}\\
  unopened and absent\\ from the corpus};

\draw[arrow] (traces) -- (occ);
\draw[arrow] (occ) -- (match);
\draw[arrow] (occ) -- (diag);
\draw[arrow] (match) -- (qa);
\draw[arrow] (diag) -- (qa);
\draw[holdout] (traces.south) -- node[right, font=\tiny, text=CPAgray]
  {held out} (sealed.north);
\end{tikzpicture}}
\caption{\textbf{Production accounting under frozen v1.0.} All 244 expected,
non-sealed trajectories completed.  The 20 sealed evaluation trajectories were
not opened.  Diagnostic records did not modify the frozen library.}
\label{fig:production-accounting}
\end{figure}
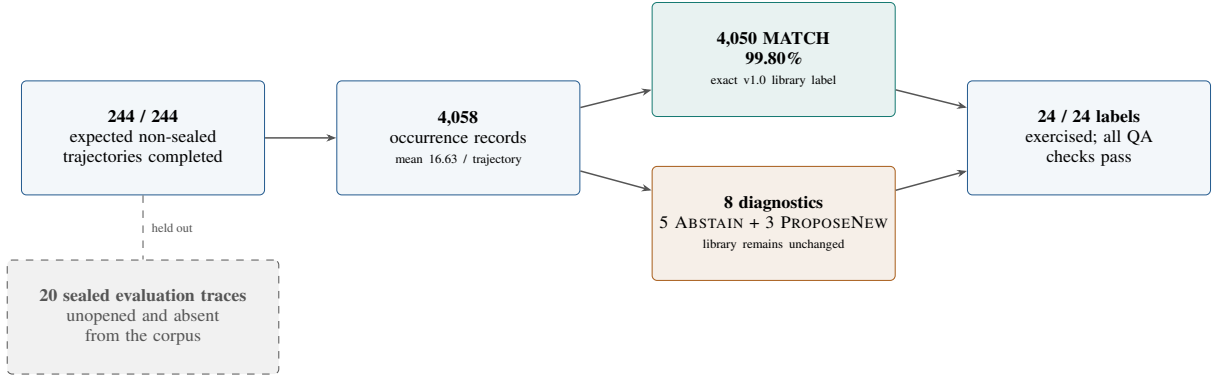

The production application is a frozen-v1.0 pass over the 244 non-sealed
trajectories.  Figure~\ref{fig:production-accounting} separates its 4,050
frozen-library matches from five abstentions and three proposed-label
diagnostics.  The latter remain in the record precisely so that a subsequent
version can reconsider them without rewriting v1.0.

\begin{table}[t]
\centering
\small
\caption{Production checks and their evidential scope.}
\label{tab:production-qa}
\begin{tabularx}{\linewidth}{p{0.42\linewidth}Y}
\toprule
Observed result & Interpretation \\
\midrule
244/244 non-sealed trajectories completed; 4,058 records & Completeness of the
planned production pass. \\
4,050 \match{}; 5 \abstain{}; 3 \propose{} diagnostics & Frozen-library
accounting; the diagnostics remain visible for future versioning. \\
All 24 frozen labels occur & The production pass exercises every dictionary
entry, including low-frequency procedures. \\
Parser, vocabulary, ID, action/context, uniqueness, ordering, and
sequence-field checks pass & Mechanical conformance to the frozen occurrence
contract. \\
245 occurrences contain interleaved non-agent events; 487 events moved to
context & Normalization preserves agent-only action lists without splitting or
collapsing an occurrence. \\
20 sealed trajectories absent and unopened & The declared evaluation boundary
is preserved. \\
\bottomrule
\end{tabularx}
\end{table}

\section{Frozen v1.0 CPA dictionary}
\label{app:dictionary}

This is the complete 24-entry frozen v1.0 library used in production.  The
table preserves each entry's operational definition and gives its
\match{}-record frequency; its counts sum to 4,050, excluding five abstentions
and three proposed-label diagnostics.  They are descriptive rather than
label-quality scores.  The accompanying source bundle provides the
machine-readable entry definitions, including inclusion/exclusion criteria
and versioned provenance (Appendix~\ref{app:repro}).

\begingroup
\renewcommand{\texttt}[1]{{\ttfamily\scriptsize #1}}
\setlength{\tabcolsep}{3pt}
\begin{longtable}{p{0.43\linewidth}p{0.48\linewidth}r}
\caption{Frozen $\tau^3$-Retail CPA dictionary and \match{}-record counts.}
\label{tab:dictionary}\\
\toprule
Label & Operational definition & $n$ \\
\midrule
\endfirsthead
\toprule
Label & Operational definition & $n$ \\
\midrule
\endhead
\path{ACKNOWLEDGE_MUTATION_COMMIT} & Report a committed mutation and its
consequences, closing the request thread.  Requires a submitted mutation with
an observed result; it is not a pre-commit proposal or bare courtesy. & 267 \\
\path{ANSWER_CUSTOMER_QUERY} & Answer an explicit customer question with
a value read from a retrieved record or derived from one non-monetarily. & 129 \\
\path{BIND_PARAMETER_FROM_RECORD} & Fill a required argument from an
already retrieved record instead of asking the customer for that value. & 175 \\
\path{COMPUTE_SETTLEMENT_AMOUNT} & Derive a monetary quantity that no
single retrieved record contains. & 248 \\
\path{DECLARE_INFORMATION_UNAVAILABLE} & Close an information request by
declaring that the requested datum cannot be obtained from available records
and tools, and name why. & 13 \\
\path{DECLINE_OUT_OF_SCOPE_REQUEST} & Refuse a request because it lies
outside what the agent is permitted or able to do, returning responsibility for
it to the customer. & 38 \\
\path{ELICIT_GOAL_SELECTION} & Name mutually exclusive service goals and
ask the customer to choose one before any goal has been prepared as an
operation. & 15 \\
\path{EXECUTE_AUTHORIZED_MUTATION} & Use an agent tool event to write an
authorized change to an order, reservation, or account profile.  Endpoint and
mutation object are occurrence attributes, not separate CPAs. & 368 \\
\path{OFFER_ALTERNATIVE_COURSES} & After the requested route is found
infeasible, propose admissible alternatives and put the choice to the customer.
& 60 \\
\path{PRESENT_VARIANT_OPTIONS} & Externalize the admissible option space
so that the customer can choose within it. & 97 \\
\path{REQUEST_IDENTITY_CREDENTIAL} & Ask the customer for a value by
which their account can be located; the customer's reply is contextual evidence,
not part of the action. & 399 \\
\path{RESOLVE_CUSTOMER_IDENTITY} & Submit supplied identifying values to
the account directory and obtain an account identifier. & 268 \\
\path{RESOLVE_REQUEST_REFERENT} & Bind an informal customer description
to the concrete record entity that it denotes. & 285 \\
\path{RETRIEVE_ACCOUNT_PROFILE} & Fetch the account record carrying
addresses, payment instruments, and the order index. & 241 \\
\path{RETRIEVE_ORDER_RECORD} & Obtain order records from the customer's
account, using one or more agent tool events when jointly needed. & 293 \\
\path{RETRIEVE_PRODUCT_CATALOG_INDEX} & Fetch the catalogue-wide
product/category index so a product named only in words can be located as an
identifier. & 16 \\
\path{RETRIEVE_PRODUCT_VARIANT_SET} & Obtain one or more products'
variant space and prices. & 184 \\
\path{SCREEN_CANDIDATE_SATISFACTION} & Issue a satisfied/not-satisfied
verdict about whether a retrieved candidate set contains something meeting the
customer's constraints. & 81 \\
\path{SCREEN_ROUTE_ADMISSIBILITY} & Issue a permit/block verdict on a
requested operation, grounded in the target record's own state. & 152 \\
\path{SELECT_VARIANT_MEETING_CONSTRAINTS} & Name the single variant
that satisfies the customer's stated criteria. & 119 \\
\path{SEQUENCE_MUTATION_STEPS} & State an order over already specified
operations on the same artifact when a known constraint makes the order
material. & 12 \\
\path{SOLICIT_MUTATION_INPUT} & Put a pending mutation to the customer and
obtain what it still needs to advance: a value, choice, ratification, or
authorization to execute. & 499 \\
\path{SURVEY_ORDER_PORTFOLIO} & Present the customer's orders as a set
with statuses before a single order has been bound as the target. & 75 \\
\path{TRANSFER_TO_HUMAN_AGENT} & Hand the unresolvable remainder of the
episode to a human agent and release control. & 16 \\
\bottomrule
\end{longtable}
\endgroup

\subsection{Dictionary-wide interpretation rules}

All entries are matched through the four fields in Eq.~\ref{eq:identity}:
proximal goal, principal output, procedural state transition, and
action-defining preconditions.  The following are explicitly non-diagnostic for
CPA identity: tool endpoint, identifier channel, record source, scenario,
mutation object, wording, and phase.  For example, different mutation endpoints
remain \path{EXECUTE_AUTHORIZED_MUTATION} when their procedural function
is the same; a catalogue index and a variant set remain different because their
goals and outputs differ.

The library also carries two cross-cutting safeguards.  First, policy-rule
clauses are contextual evidence, not independent CPA occurrences.  Second,
\path{SOLICIT_MUTATION_INPUT} is deliberately an observation-level merge
of conceptual parameter elicitation and consent/ratification when the needed
intermediate selection state is not observable.  It must not be read as a claim
that those latent activities are identical.  This documented breadth is why its
definition contains multiple permitted input forms and why the R6 audit treats
it as a dedicated stability check.

\section{Retrospective validation and outstanding experiments}
\label{app:validation}

These analyses were added after the original freeze.  They use the original
outputs and do not revise v1.0.  Human validity, finer semantic annotation,
downstream utility, and model/domain replication are proposed follow-up
studies, with no results imputed for them.

\subsection{Per-label evidence and disagreement cases}

\begingroup
\scriptsize
\setlength{\tabcolsep}{3pt}
\begin{longtable}{p{.47\linewidth}rrrrrr}
\caption{All 24 labels: R6 counts per context, matched multiplicity $a_k$,
agreement $A(k)$, distinct R6 traces $T_k$, and production matches $N_{\rm prod}$.
A dash means no R6 observations, not perfect agreement.}\label{tab:per-label}\\
\toprule Label & C1 & C2 & $a_k$ & $A(k)$ & $T_k$ & $N_{\rm prod}$\\\midrule\endfirsthead
\toprule Label & C1 & C2 & $a_k$ & $A(k)$ & $T_k$ & $N_{\rm prod}$\\\midrule\endhead
\path{ACKNOWLEDGE_MUTATION_COMMIT} & 42 & 42 & 42 & 1.000 & 30 & 267 \\
\path{ANSWER_CUSTOMER_QUERY} & 8 & 8 & 8 & 1.000 & 8 & 129 \\
\path{BIND_PARAMETER_FROM_RECORD} & 27 & 22 & 20 & 0.816 & 26 & 175 \\
\path{COMPUTE_SETTLEMENT_AMOUNT} & 19 & 19 & 18 & 0.947 & 16 & 248 \\
\path{DECLARE_INFORMATION_UNAVAILABLE} & 2 & 2 & 2 & 1.000 & 2 & 13 \\
\path{DECLINE_OUT_OF_SCOPE_REQUEST} & 6 & 6 & 5 & 0.833 & 5 & 38 \\
\path{ELICIT_GOAL_SELECTION} & 1 & 1 & 1 & 1.000 & 1 & 15 \\
\path{EXECUTE_AUTHORIZED_MUTATION} & 55 & 55 & 55 & 1.000 & 30 & 368 \\
\path{OFFER_ALTERNATIVE_COURSES} & 7 & 7 & 7 & 1.000 & 5 & 60 \\
\path{PRESENT_VARIANT_OPTIONS} & 6 & 6 & 6 & 1.000 & 6 & 97 \\
\path{REQUEST_IDENTITY_CREDENTIAL} & 51 & 51 & 51 & 1.000 & 31 & 399 \\
\path{RESOLVE_CUSTOMER_IDENTITY} & 34 & 34 & 34 & 1.000 & 32 & 268 \\
\path{RESOLVE_REQUEST_REFERENT} & 32 & 29 & 29 & 0.951 & 24 & 285 \\
\path{RETRIEVE_ACCOUNT_PROFILE} & 33 & 33 & 33 & 1.000 & 32 & 241 \\
\path{RETRIEVE_ORDER_RECORD} & 39 & 39 & 39 & 1.000 & 32 & 293 \\
\path{RETRIEVE_PRODUCT_CATALOG_INDEX} & 1 & 1 & 1 & 1.000 & 1 & 16 \\
\path{RETRIEVE_PRODUCT_VARIANT_SET} & 20 & 20 & 20 & 1.000 & 18 & 184 \\
\path{SCREEN_CANDIDATE_SATISFACTION} & 7 & 7 & 7 & 1.000 & 7 & 81 \\
\path{SCREEN_ROUTE_ADMISSIBILITY} & 20 & 19 & 19 & 0.974 & 18 & 152 \\
\path{SELECT_VARIANT_MEETING_CONSTRAINTS} & 12 & 12 & 12 & 1.000 & 11 & 119 \\
\path{SEQUENCE_MUTATION_STEPS} & 0 & 0 & 0 & --- & 0 & 12 \\
\path{SOLICIT_MUTATION_INPUT} & 72 & 72 & 72 & 1.000 & 32 & 499 \\
\path{SURVEY_ORDER_PORTFOLIO} & 4 & 5 & 4 & 0.889 & 5 & 75 \\
\path{TRANSFER_TO_HUMAN_AGENT} & 1 & 1 & 1 & 1.000 & 1 & 16 \\
\bottomrule
\end{longtable}

\endgroup

Counts in Table~\ref{tab:per-label} refer to occurrence multiplicities at
trace--anchor--label keys, not unique label strings.  The 23-label macro mean
is \CPAMacroAgreement.  The three single-occurrence entries each occur in
only one trajectory; their perfect scores are not precise estimates.
Production counts show exposure, not accuracy, and do not fill the missing
R6 evidence for \path{SEQUENCE_MUTATION_STEPS}.

The 18 unmatched occurrences lie at 16 anchors in 12 traces.  We retain the
two residual label multisets after subtracting exact matches, rather than
forcing extra occurrences into a one-to-one confusion matrix.  A discrepancy
can be an extra action, a missing action, a different label, or a shifted
anchor.  Neither context is designated as correct.  Three examples expose
the questions requiring adjudication:

\begin{itemize}
  \item At T101-0, event 12, C1 has an unmatched
  \path{SCREEN_ROUTE_ADMISSIBILITY} and C2 an unmatched
  \path{DECLINE_OUT_OF_SCOPE_REQUEST}.  The message explains limits
  on changing a product type while asking for modification details.
  Is this an operation-class refusal or a route-admissibility verdict?
  \item At T113-2, event 20, C1 additionally records
  \path{BIND_PARAMETER_FROM_RECORD} when the agent names the original
  payment method in a cancellation confirmation.  C2 does not.  Does that
  clause separately bind a value or restate an already bound value?
  \item In T60-3, C1 has an additional
  \path{COMPUTE_SETTLEMENT_AMOUNT} at event 10 and C2 at event 12.
  Both messages state price differences.  Aggregate label totals conceal
  this disagreement about which event first realizes the action.
\end{itemize}

These are disagreement diagnoses, not adjudicated errors.  The analysis
artifact retains all disputed anchors and their visible messages.  The
additional evidence-reference comparison uses the same 499/491 rows, with
keys $(t,h,\ell,\operatorname{set}(\mathbf x))$.  Matching is by multiset
intersection, giving 395 matches, 104 C1-only rows, and 96 C2-only rows.
Adding the stored span endpoint leaves these counts unchanged.  Context
list ordering and duplication do not change a reference set; repeated
occurrence records still count separately.

\subsection{What the granularity comparison measures}

For a deterministic many-to-one map $f$ on labels, define
$n^{(r)}_{thg}=\sum_{k:f(k)=g}n^{(r)}_{thk}$.
Since
\begin{equation}
\min\!\left\{\sum_k x_k,\sum_k y_k\right\}
\ \geq\ \sum_k\min\{x_k,y_k\},
\end{equation}
coarsening cannot decrease aggregate overlap when anchors and occurrences are
held fixed.  In the one-label case, $A_f=F_{\rm mult}$ exactly.  Higher
agreement after merging is therefore not independent evidence for the merger.

Our eight-family mapping groups identity requests/resolution; record and
catalog retrieval; referent resolution, portfolio survey, and parameter
binding; screening, constrained selection, out-of-scope refusal, and mutation
ordering; goal selection and mutation-input solicitation; mutation execution;
acknowledgment, answers, amounts, unavailability, alternatives, and variant
presentation; and human handoff.  The analysis artifact enumerates the map.
It is an illustrative post-hoc coarsening, not a validated ontology.  Its
agreement is \CPACoarseAgreement, compared with 0.982 for 24 labels and
0.986 for one label.  It eliminates two same-anchor cross-label residual
pairs, leaving the multiplicity discrepancy.

A substantive fine-grained comparison requires new independent coding.
One proposed 27-entry candidate replaces
\path{SOLICIT_MUTATION_INPUT} with missing-value elicitation,
choice among options, ratification of a proposed value, and final execution
authorization.  Definitions must specify observable distinctions and retain
abstention where the distinctions are unsupported.  Separate procedural
outputs may still yield co-anchored occurrences.  This candidate has not been
applied here.  A future study should compare 8, 24, and 27 entries on the same
task-disjoint traces, report multiplicity and per-label stability, and test
whether the extra distinctions improve diagnosis of authorization or
parameter-binding failures.  Revising a codebook after inspecting evaluation
disagreements would require a new evaluation sample.

\subsection{Native trace and deterministic-rule comparisons}

\begin{table}[ht]
\centering
\scriptsize
\begin{tabular}{lrrr}
\toprule Representation & Records & $F_{\rm mult}$ & Label overlap\\\midrule
Native agent events (one per event) & 426 & 0.765/0.772 & --- \\
Endpoint rule (one per call) & 251 & 0.493/0.499 & 0.493/0.499 \\
Endpoint rule (group repeated retrievals) & 186 & 0.540/0.547 & 0.540/0.547 \\
\bottomrule
\end{tabular}

\caption{Exploratory reconstruction baselines on R6.  Paired scores compare
each baseline separately with C1/C2.  Label overlap uses the same multiset
formula as $A$; these LLM references are not gold annotations.  The native
baseline retains event types rather than predicting CPA labels.}
\label{tab:baselines}
\end{table}

The native baseline includes all assistant-message and tool-call events except
the initial greeting, including later courtesy messages.  The endpoint map
sends user-ID lookups to identity resolution, user/order-detail tools to
their retrieval labels, product/item-detail tools to variant-set retrieval,
product-type listing to catalog-index retrieval, the calculator to settlement
computation, mutation endpoints to authorized mutation execution, and the
handoff endpoint to human transfer.  The word ``authorized'' in that mapped
label does not verify consent: the rule examines only the endpoint.

There are 251 raw tool calls.  Grouping consecutive retrievals leaves 186
rule occurrences, of which 185 match each context.  On tool anchors alone,
one-call and grouped-rule overlap are 0.849 and 0.997.  The grouped rule's
precision against each full annotation is 0.995, while coverage is only
0.371/0.377 because it emits no conversational actions.  These numbers
identify an additional annotation burden; they do not establish its utility.

A downstream test should compare native traces alone, native traces plus
deterministic labels, and native traces plus CPA, using the same complete
source evidence.  Include a development-tuned message-cue rule baseline as
well as endpoint rules.  Freeze rules, failure definitions, and time budgets
before held-out scoring; use task-disjoint development/evaluation groups,
counterbalance conditions, and prevent evaluators from seeing the same case
in multiple conditions.  Independent adjudicators must establish failure
presence and location from raw evidence before viewing the experimental
labels.  Report diagnosis precision/recall, localization, evidence
sufficiency, time per case, and annotation cost, with task-cluster uncertainty.
No downstream measurements are available in this paper.

\subsection{Independent human reference: prepared, not executed}

Two annotators should receive complete event-indexed traces and the frozen
dictionary, with neither LLM suggestions nor each other's labels.  This avoids
turning agreement with a suggestion into an apparent independent reference
\citep{schroeder2025anchoring}.  First enumerate actions and their observable
outputs, then assign labels and cite supporting events.  Include all agent
events: reviewing only the union of model outputs cannot reveal shared
omissions.  Distinguish valid reference IDs, sufficient visible support, and
support available only in a full payload absent from the historical rendering.

The prepared retrospective pilot uses the 32 R6 traces and a separate targeted
production supplement for rare labels and diagnostic decisions.  Selecting
with existing LLM labels can miss rare actions omitted by the model; the
supplement cannot estimate prevalence or corpus-wide accuracy.  Preserve both
independent human passes before adjudication by a third person.  Adjudication
separates omissions, unsupported additions, label errors, multiplicity/anchor
errors, and insufficient evidence, and may leave cases unresolved.  Model
majority voting is not a substitute.  Report human--human agreement before
adjudication and model--reference action and label precision/recall afterwards,
alongside evidence sufficiency and unresolved cases.  Analyze the R6 pilot
and targeted supplement separately.  A confirmatory accuracy claim requires a
newly reserved task-disjoint sample, sized for a declared precision before
its labels are seen.

\subsection{Replication and transfer: preparations are not results}

A new run must record the exact provider/model identifier or explicitly mark
it unavailable, harness version, decoding parameters, context limit, prompt
and dictionary hashes, rendering policy, task IDs, retry history, and output
hashes.  Use at least three fresh runs per model on the same locked retail
sample, with at least two model systems; report within-model and between-model
overlap separately.  The historical model change is confounded with shard
assignment and cannot serve as a cross-model experiment.

For a second domain, distinguish applying the retail library unchanged from
rebuilding a domain-specific library with the CPA protocol.  The former
tests coverage and must retain proposals and abstentions; the latter tests
the construction procedure and needs its own development and untouched audit
tasks.  Existing local airline/telecom materials are preparation artifacts,
not completed validation evidence for this paper.  Missing decoding settings
and unavailable deployment identifiers must remain explicit.

\section{Artifacts, provenance, and reproducibility limits}
\label{app:repro}

The accompanying source package places the numerical reproduction bundle
under \path{anc/}.  It contains hash-checked historical inputs, both original
R6 response pairs, normalized occurrences, production responses and rendered
packs, the codebook, and source manifests.  It does not require a live model
service or access to the original Git repository.  Historical repository paths
and commit identifiers are provenance metadata.  Two historical scripts have
machine-specific raw-file paths replaced by a neutral filename in the public
bundle; its manifest retains both original and packaged hashes and records
these redactions.  Annotation data are unchanged.

\begin{longtable}{p{0.44\linewidth}p{0.48\linewidth}}
\caption{Paths within the accompanying analysis bundle (under \texttt{anc/}).}
\label{tab:artifact-map}\\
\toprule Artifact & Role \\
\midrule
\endfirsthead
\toprule Artifact & Role \\
\midrule
\endhead
\path{analysis/INPUT_MANIFEST.json} & Packaged hashes and provenance of 67 historical inputs; original hashes retained for the two path-redacted scripts. \\
\path{analysis/inputs/library/} & Frozen v1.0 dictionary, manifest, and freeze report. \\
\path{analysis/inputs/development/} & Archived candidate libraries, schema, and consolidation report. \\
\path{analysis/inputs/R1/} through \path{analysis/inputs/R6/} & Sampling records; available audit provenance; original R6 packs, responses, occurrences, and freeze decision. \\
\path{analysis/inputs/production/} & All 16 production response files and rendered packs, sequence corpus, QA summary, and shard manifest. \\
\path{analysis/analyze.py} & Historical reconstruction, original overlap scores, stratification, coarsening, baselines, and task-cluster intervals. \\
\path{analysis/query_demo.py} & Co-anchored action query and exact context-reference overlap. \\
\path{analysis/outputs/} & Machine-readable numerical results, mappings, disagreement examples, and query results. \\
\path{analysis/test_analysis.py} and \path{analysis/test_reference_overlap.py} & Metric boundary, reference-set, and multiplicity checks. \\
\bottomrule
\end{longtable}

\subsection{Offline reproduction}

From the bundle root, with Python 3.8 or later:
\begin{lstlisting}
python3 -B analysis/analyze.py
python3 -B analysis/query_demo.py
python3 -B -m unittest discover -s analysis -p 'test_*.py'
\end{lstlisting}
The first command verifies the archived input hashes, reparses the raw R6
responses and checks their equality to all 990 archived rows, reconstructs all
4,058 production records, and checks trace separation and reserve exclusion.
It emits the historical analyses and generated tables.  The second emits the
matching trace/event IDs for the query in Section~\ref{sec:query-example},
computes the reference-set comparisons, and generates their manuscript macros.
The third checks seven metric boundary cases.  Generated tables can then be
compared with the copies under the manuscript's \path{generated/} directory.

The bundle recomputes the reported R6 and production numbers, but does not
replay all historical development runs.  The prepared human pilot has no
collected responses and supplies no validation evidence.

\subsection{Evidence and configuration limits}

The production sequence file contains action-event IDs but omits context-event
lists; these references require the retained raw responses and source packs.
The sequence file alone is therefore not a complete evidence archive.
Historical R6 arguments and tool results were clipped to 110 characters.
Full payloads in any future human study must be distinguished from the
information available to the historical annotators.

The R6 provenance manifest declares two isolated Claude Code annotation
contexts, \texttt{claude-opus-5}, and a one-million-token context window, with
no cross-visibility.  Temperature and top-$p$ were not pinned.  Production
used one annotation stream with a configuration change: shards s03--s04 used
the earlier configuration, while s01--s02 and s05--s16 used the later one.
These are historical declarations, not independently verified deployment
snapshots; the change is confounded with shard assignment.  It is not a
controlled cross-model experiment.

The exact benchmark revision and trace-generating agent configuration remain
unrecovered; benchmark citations do not resolve that provenance.  Identical
new annotations and trajectory regeneration are not established.  No human
reference, cross-domain result, or utility measurement is included.  Reuse
these annotations as inspectable model outputs, not ground truth.

\end{document}